\documentclass[acmtog,nonacm]{acmart} 
 
\usepackage{booktabs} 

\usepackage[ruled]{algorithm2e} 

\SetAlFnt{\small}
\SetAlCapFnt{\small}
\SetAlCapNameFnt{\small}
\SetAlCapHSkip{0pt}

\setcopyright{acmcopyright}\acmJournal{TOG}
\acmYear{2022}\acmVolume{41}\acmNumber{4}\acmArticle{99}\acmMonth{7} \acmDOI{10.1145/3528223.3530122}

\usepackage{hyperref}
\usepackage{url}
\usepackage{pgfplots}
\pgfplotsset{compat=1.15}
\usepackage{siunitx}
\usepackage{xcolor}

\graphicspath{{images/}}

\def\authornote#1#2#3{{\textcolor{#2}{\textsl{\small#1:[*#3*]}}}}

\newcommand{\DR}[1]{}
\newcommand{\LF}[1]{}
\newcommand{\MS}[1]{}

\begin{document}
\title{ADOP: Approximate Differentiable One-Pixel Point Rendering}

\author{Darius R\"uckert}
\email{darius.rueckert@fau.de}
\affiliation{%
	\institution{Friedrich-Alexander-Universit\"at Erlangen-N\"urnberg}
	\city{Erlangen}
	\country{Germany}
}

\author{Linus Franke}
\email{linus.franke@fau.de}

\author{Marc Stamminger}
\email{marc.stamminger@fau.de}
\affiliation{%
	\institution{Friedrich-Alexander-Universit\"at Erlangen-N\"urnberg}
	\city{Erlangen}
	\country{Germany}
}

\renewcommand\shortauthors{R\"uckert et al.}



\begin{teaserfigure}
	\centering
	\includegraphics[width=.9\linewidth]{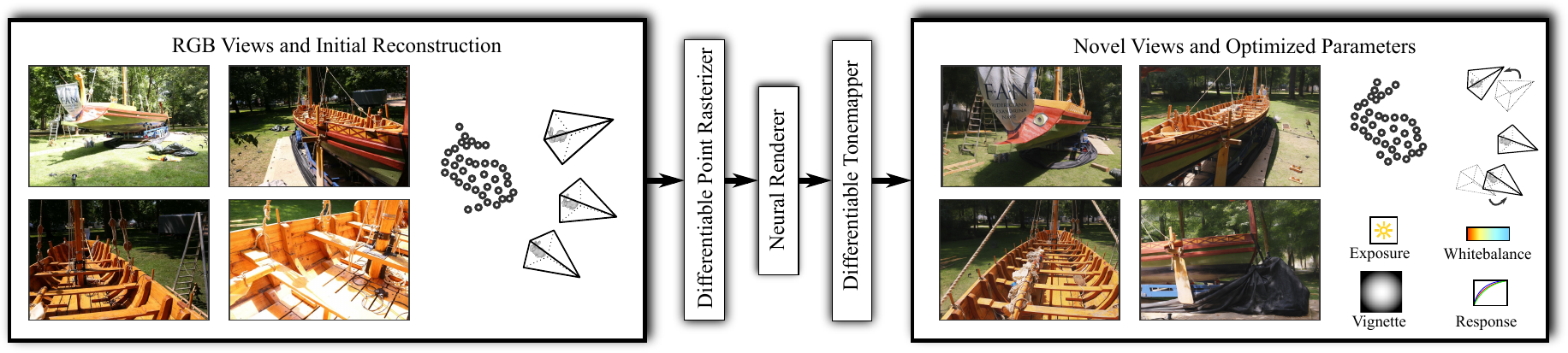}%
	\centering
	\caption{Given a set of RGB images and an initial 3D reconstruction (left), our inverse rendering approach is able to synthesize novel frames and optimize the scene's parameters (right), for instance point position and camera pose as well as image settings such as exposure time and white balance. In combination with a fast point-based renderer, we achieve real-time render times even for complex environments.}
	\label{fig:teaser}
\end{teaserfigure}

\begin{abstract}
	In this paper we present ADOP, a novel point-based, differentiable neural rendering pipeline.
	Like other neural renderers, our system takes as input calibrated camera images and a proxy geometry of the scene, in our case a point cloud.
	To generate a novel view, the point cloud is rasterized with learned feature vectors as colors and a deep neural network fills the remaining holes and shades each output pixel.
	The rasterizer renders points as one-pixel splats, which makes it very fast and allows us to compute gradients with respect to all relevant input parameters efficiently.
	Furthermore, our pipeline contains a fully differentiable physically-based photometric camera model, including exposure, white balance, and a camera response function.
	Following the idea of inverse rendering, we use our renderer to refine its input in order to reduce inconsistencies and optimize the quality of its output.
	In particular, we can optimize structural parameters like the camera pose, lens distortions, point positions and features, and a neural environment map, but also photometric parameters like camera response function, vignetting, and per-image exposure and white balance.
	Because our pipeline includes photometric parameters, e.g.~exposure and camera response function, our system can smoothly handle input images with varying exposure and white balance, and generates high-dynamic range output.
	We show that due to the improved input, we can achieve high render quality, also for difficult input, e.g. with imperfect camera calibrations, inaccurate proxy geometry, or varying exposure.
	As a result, a simpler and thus faster deep neural network is sufficient for reconstruction.
	In combination with the fast point rasterization, ADOP achieves real-time rendering rates even for models with well over 100M points.
	\begin{center}
		{\color{blue}\url{https://github.com/darglein/ADOP}}
	\end{center}
\end{abstract}

%
%

%
%


\maketitle

\section{Introduction}
\label{sec:intro}

With neural rendering it is possible to generate stunning free-viewpoint reconstructions of real world scenes.
The field has advanced enormously in the last years with two major streams emerging~\cite{tewari2020state}: Proxy-based approaches that use some 3D proxy to carry reconstructed scene information (e.g., triangle meshes \cite{thies2019deferred} or point clouds \cite{aliev2020neural}), and direct approaches that encode light or radiance fields (e.g., NeRFs \cite{mildenhall2020nerf}).

Approaches with geometric proxies extend the idea of previous image based rendering methods \cite{chaurasia2013depth}, with the difference that neural networks are used to interpolate and blend the input images on the geometric proxy.
The rationale is that neural networks can achieve a better interpolation of view-dependent effects as well as handle imperfections of the geometric proxy and the resulting reprojection errors.
Although this holds true to some degree, inconsistencies in the input still have significant impact on the quality of the result.
As such, slight miscalibration of intrinsic and extrinsic camera parameters can lead to blurred results.
Even worse, varying camera exposure in the input images cannot be compensated well and leads to patches of inconsistent brightness.

In this paper, we present a novel neural rendering system that directly accounts for such inconsistencies and renders very efficiently.
As input we assume a number of images of a scene.
A standard 3D reconstruction pipeline estimates intrinsic and extrinsic camera parameters and reconstructs a point cloud of the scene geometry.
Following prior point-based neural rendering work \cite{aliev2020neural}, we learn feature vectors for all points and generate novel views by rendering the point cloud to an image pyramid, from which a deep neural network generates the final image.

However, our renderer is differentiable with respect to structural input parameters, such as intrinsic and extrinsic camera parameters, camera exposure, and even point cloud positions.
The core idea is an approximation of the spatial gradients of one-pixel point splatting, an otherwise non-differentiable operation.
We thus call our method ADOP: Approximate differentiable one-pixel point rendering.
Using this technique, we can apply the idea of inverse rendering and use the renderer itself to improve its output by optimizing the input (see Fig.~\ref{fig:teaser}).
In this paper, we show that this self-calibration reduces a large variety of sources of inconsistencies in the input, resulting in high-quality novel view synthesis, less constraints on the input, and faster rendering times.

In detail, ADOP makes the following contributions:
\begin{itemize}
\item We present a point-based rendering pipeline, that can self-optimize all structural input parameters obtained from the reconstruction, namely camera poses, lens distortions, and even point cloud positions, which leads to significantly improved render quality.
\item We show how to optimize for camera response and vignetting, as well as per-image exposure and white balancing to support input images with varying exposure, as it is almost unavoidable for outdoor scenes.
As a side effect, the learned features encode the scene texture in high-dynamic range, and it is possible to apply arbitrary tone mapping operators to the synthesized views.
\item We present a highly efficient rasterizer and rendering optimizations with which our system achieves real-time rendering rates.
We show that due to the significantly improved input, smaller networks are sufficient to achieve good rendering quality.
Together with a highly optimized point renderer, we can render photo realistic full-HD images in 30FPS on off-the-shelf hardware.
\end{itemize}

\begin{figure*}
	\includegraphics[width=\linewidth]{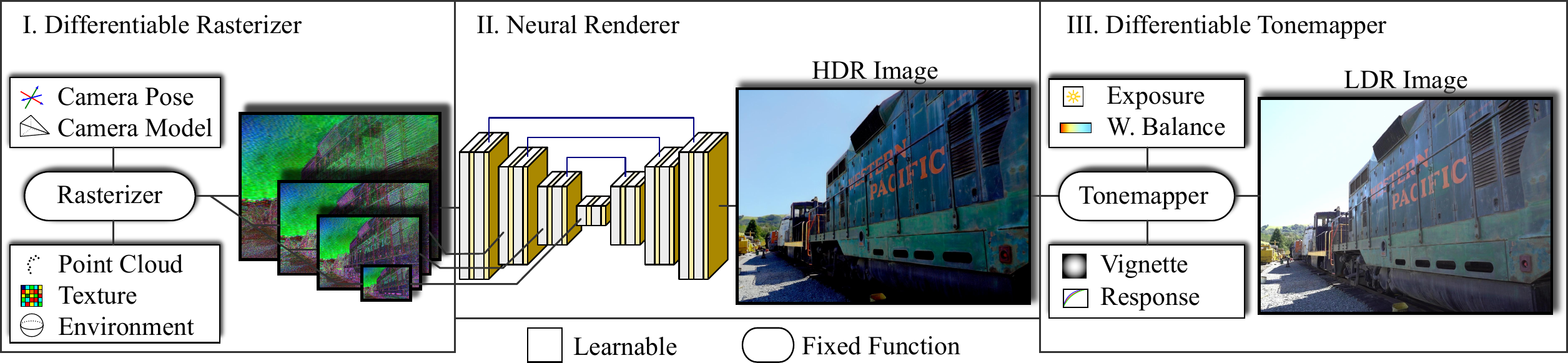}
	\caption{
	Overview of our point-based HDR neural rendering pipeline. 
	The scene, consisting of a textured point cloud and an environment map, is rasterized into a set of sparse neural images in multiple resolutions.
	A deep neural network reconstructs an HDR image, which is then converted to LDR by a differentiable physically-based tonemapper.
	All parameters in the rectangular boxes, as well as the neural network can be optimized simultaneously during training.	
	}
	\label{fig:pipeline_overview}
\end{figure*}

\section{Related Work}
\label{sec:relatedwork}

\subsection{Novel View Synthesis}
Traditional novel view synthesis, which is closely related to image-based rendering (IBR), relies on the basic principle of warping colors from one frame to another.
One approach is to use a triangle-mesh proxy to directly warp the image colors to a novel view~\cite{debevec1998efficient,chaurasia2013depth,shum2000review}.
However, imperfect geometry leads to ghosting artifacts around silhouettes.
This can be improved by replacing hand-crafted heuristics in the classic IBR pipeline with deep neural networks~\cite{hedman2018deep,riegler2020free,riegler2021stable,thies2020image}.
If no proxy geometry is available, learning-based approaches have been developed that create a multi plane image representation~\cite{srinivasan2019pushing,tucker2020single,zhou2018stereo,mildenhall2019local} or directly estimate the required warp-field~\cite{ganin2016deepwarp,zhou2016view,flynn2016deepstereo}.
Novel view synthesis can also be performed by reconstructing a 3D model of the scene and rendering it from novel view points.
\citet{thies2019deferred} learn a neural texture on a triangle mesh which can be rendered using traditional rasterization.
The rasterized image is then converted to RGB by a deep neural network.
Other approaches use ray-casting to automatically learn a voxel grid~\cite{nguyen2018rendernet,sitzmann2019deepvoxels,zhu2018visual} or an implicit function~\cite{park2019deepsdf,mescheder2019occupancy}.
It has also been shown that point clouds are suitable geometric proxies for novel view synthesis~\cite{Pittaluga_2019_CVPR,meshry2019neural}.
\textit{Neural Point-based Graphics} (NPBG)~\cite{aliev2020neural}, which is closely related to our method, renders a point cloud with learned neural descriptors in multiple resolutions.
These images are then used to reconstruct the final output image by a deep neural network.
NPBG is very flexible as it doesn't require a textured triangle mesh as proxy and shows impressive results.
Similarly, \citet{Dai_2020_CVPR} show that rendering multiple depth layers of a point cloud can also be used to synthesize novel views.

\subsection{Inverse Rendering}
Inverse rendering and differentiable rendering have been a topic of research for some time. 
However, major breakthroughs have only been made in recent years due to improved hardware and advancements in deep learning~\cite{kato2020differentiable}.
Its application can be challenging though:
Traditional triangle rasterization with depth-testing has no analytically correct spatial derivative~\cite{loper2014opendr,kato2020differentiable}.
Available systems therefore either approximate the gradient~\cite{loper2014opendr,kato2018neural,kato2019learning,genova2018unsupervised} or approximate the rendering itself using alpha blending along edges~\cite{rhodin2015versatile,liu2019soft,chen2019learning}.
Volume raycasting on the other hand is differentiable by accumulating all voxel values along the ray ~\cite{kato2020differentiable}.
This has been used by multiple authors to build volumetric reconstruction systems~\cite{tulsiani2017multi,henzler2019escaping,lombardi2019neural} or predict the signed distance field of an object~\cite{jiang2020sdfdiff}.
Instead of a voxel grid, an implicit function can also be used inside a differentiable volumetric raycasting framework~\cite{liu2019learning,niemeyer2020differentiable,liu2020dist,zakharov2020autolabeling}.
\citet{mildenhall2020nerf} show with \textit{Neural Radiance Fields} (NeRF) that a deep neural network can be trained by volumetric raycasting to store the view-dependent radiance of an object.
Due to the impressive results of NeRF, multiple extensions and improvements have been published in the following years~\cite{zhang2020nerfpp,yu2021pixelnerf,martin2021nerf,sitzmann2020implicit}.
Inverse rendering has also been proposed for point cloud rasterization~\cite{kato2020differentiable,lin2018learning}.
The spatial gradients of the points can be approximated in different ways.
This includes gradient computation along silhouettes~\cite{han2020drwr}, gradients for Gaussian splatting~\cite{yifan2019differentiable,wiles2020synsin,kopanas2021point}, and gradient approximation using a temporary 3D volume~\cite{insafutdinov2018unsupervised}.
A different approach is taken by Lassner and Zollh\"ofer~\shortcite{Lassner_2021_CVPR} who render the points as small spheres instead of splats. 
Our differentiable point rendering approach is similar to established differentiable splatting techniques~\cite{yifan2019differentiable,kopanas2021point},
however we render only one pixel per point, which allows our approach to be multiple magnitudes more efficient than competing methods~\cite{Lassner_2021_CVPR}.
A nice property of a differentiable rendering pipeline is that the camera parameters can be optimized during rendering.
\citet{lin2021barf} and \citet{jeong2021self} show that the camera model and camera pose can be optimized in the NeRF pipeline by providing a differentiable projection module.
This allows to correct small errors in the initial calibration or even estimate the scene's geometry from scratch.
Similar work focuses on synthesizing photometric correct views by storing linear HDR radiance values inside the NeRF.
A fixed tonemapper \cite{mildenhall2021nerf} or neural tonemapper \cite{huang2021hdr} then converts the integrated HDR intensities to the LDR color space.
Concerning input refinement, this paper is inspired by these approaches and similar to concurrent work of \citet{tancik2022block}. However, we use an improved differentiable tonemapper module and apply the structural optimization on a point cloud instead of NeRFs.

\subsection{Point-Based Rendering}
Point-based rendering has been a topic of interest in computer graphics research for some time~\cite{levoy1985use}.
Over the past decades two orthogonal approaches have been developed~\cite{kobbelt2004survey}.
The first major approach is the rendering of points as oriented discs, which are usually called \textit{splats} or \textit{surfels}~\cite{zwicker2004point}, 
with the radius of each disc being precomputed from the point cloud density. 
To get rid of visible artifacts between neighboring splats, the discs are rendered with a Gaussian alpha-mask and then combined by a normalizing blend function~\cite{alexa2004point, pfister2000surfels, zwicker2001surface}.
Further extensions to surfel rendering exist, for example, Auto Splats, a technique to automatically compute oriented and sized splats~\cite{preiner2012auto}.
In recent years, deep learning-based approaches have been presented that improve image quality of surfel rendering, especially along sharp edges~\cite{bui2018point,Yang_2020_CVPR}. 

The second major approach to point-based graphics is point sample rendering~\cite{grossman1998point}, where points are rendered as one-pixel splats generating a sparse image of the scene.
Using iterative~\cite{rosenthal2008image} or pyramid-based~\cite{grossman1998point,pintus2011real,marroquim2007efficient} hole-filling approaches, the final image is reconstructed as a post processing step.
To reduce aliasing in moving scenes, points with a similar depth value can be blended during rendering~\cite{botsch2005high,schutz2021rendering}.
It has been shown that software implementations~\cite{gunther2013gpgpu,schutz2021rendering} outperform hardware accelerated rendering through \texttt{GL\_POINTS}~\cite{shreiner2009opengl}.
Recently developed approaches replace traditional hole-filling techniques with deep neural networks to reduce blurriness and better complete large holes~\cite{Pittaluga_2019_CVPR,meshry2019neural,song2020deep,le2020novel}.

\section{Neural Rendering Pipeline}
\label{sec:pipeline}

An overview of our end-to-end trainable neural rendering pipeline is shown in Fig.~\ref{fig:pipeline_overview}.
As input, we use a point cloud, which can come from common Multiview Stereo or LiDAR and SLAM systems, and the novel frame's camera parameters.
Learned neural descriptors are assigned to the point cloud.

The first step in the pipeline is a differentiable \textit{rasterizer} (Fig.~\ref{fig:pipeline_overview} left), that renders each point as a one-pixel-sized splat by projecting it to image space using the camera parameters.
%
Second, a deep \textit{neural renderer} network is used to fill in holes and shade the image.
Finally, a learnable \textit{tone mapping} operator with a camera sensor model is applied, which converts the rendered high-dynamic range (HDR) image to the displayable low dynamic range (LDR).
For optimization, input views are re-rendered, compared to the ground truth, and the loss is backpropagated through our rendering pipeline.
Since all steps are differentiable, we can simultaneously optimize all parameters, in particular the scene structure, photometric model, point features, and network parameters, which improves consistency of the input and adds robustness from a possible imperfect initial reconstruction or calibration. In the following section, we detail on these three steps.

\subsection{Differentiable One-Pixel Point Rasterization}
\label{sec:diffrast}

\subsubsection*{Forward Pass}
\label{sec:forward}
Our differentiable rasterizer renders multiple resolutions of a textured point cloud using one-pixel-sized splats.
Formally speaking, the resolution layer $l \in \{0, 1 \dots ,L-1\}$ of the neural image $I$ is the output of the render function $\Phi_l$ 
\begin{equation}
	\text{I}_l = \Phi_l(C, R, t, x,n, E, \tau),
	\label{eq:render}
\end{equation}
where $C$ is the camera model, $(R,t)$ the camera pose, $x$ the point position, $n$ the point normal, $E$ the environment map, and $\tau$ the neural texture.

The rasterizer performs three steps, which are projection, occlusion check, and blending.
The first step is the projection of each world point $x$ into the image-space of layer $l$. 
Using the camera model $C$ and the rigid transformation from world to camera-space $(R,t)$, we define this projection as:
\begin{equation}
	P_l(C,R,t,x) = \frac{1}{2^l} C(Rx +t)
\end{equation}
The real valued result of $P$ is then converted to pixel coordinates by rounding it to the nearest integer.
A world point $x_i \in \mathbb{R}^3$ is therefore projected to pixel coordinates $p_i \in \mathbb{Z}^2$ by
\begin{equation}
	p_i = \Bigl\lfloor P_l(C,R,t,x_i) \Bigr\rceil.
	\label{eq:round}
\end{equation}
Note that the rounding operation makes the projection discrete, which requires us to approximate its derivative (see later).
After all points have been projected to image space, we discard them if they fail at least one of the following conditions:
\begin{align}
\text{Bounds Test:} \quad &p_{ix} \in [0,w[\quad \land \quad p_{iy} \in [0,h[\ \label{eq:boundscheck} \\
\text{Normal Culling:}	\quad &	(Rn_i)^T \cdot \frac{Rx_i +t}{|Rx_i +t|}  > 0  \label{eq:normalculling} \\
\text{Depth Test:}	\quad &	z \le (1+\alpha) \text{min}_z(p_i)
\label{eq:depthtest}
\end{align}
The last condition \eqref{eq:depthtest} is the fuzzy depth-test as described in~\citet{schutz2021rendering}. 
A point passes the fuzzy depth test if its $z$-value is smaller than or equal to the scaled minimum depth value at that pixel.
If the camera model does not provide a valid $z$-value, we instead use the distance between the camera and the 3D point.
A large value of $\alpha$ in Eq. \eqref{eq:depthtest} increases the number of points that pass the depth test.
This leads to a smooth image but also can introduce artifacts when background points are merged into the foreground.
In practice we use $\alpha = 0.01$ as suggested by \citet{schutz2021rendering}.

\begin{figure}
	\hfil
	\includegraphics[width=.7\linewidth]{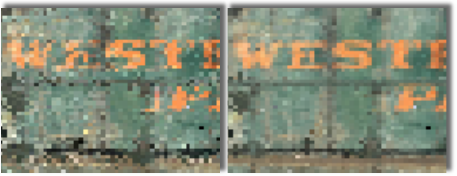}
	\caption{One-pixel point rendering with fuzzy depth testing and threshold $\alpha = 0$ on the left and $\alpha =0.01$ on the right.
	}
	\label{fig:hq_shading}
\end{figure}

After the bounds check \eqref{eq:boundscheck}, normal culling \eqref{eq:normalculling}, and fuzzy depth-test \eqref{eq:depthtest}, for each pixel $(l, u,v)$ we obtain a list of points~$\Lambda_{l,u,v}$.
If a pixel is not hit by any point, we sample the output color from the environment map $E$ using the inverse camera projection~$C^{-1}$.
Otherwise, we sample the texture $\tau$ of every point and write the mean into the output image.
The blending function $B_l$ can therefore be written as:
\begin{equation}
	 I_l(u, v, \Lambda) =
	\begin{cases}
		E(R^TC^{-1}(u,v))                  &\Lambda_{l,u,v} = \emptyset \\
		\frac{1}{|\Lambda_{l,u,v}|}\sum_{p\in \Lambda_{l,u,v}} \tau(p)                           & \text{else}
	\end{cases}
	\label{eq:blend}
\end{equation}
Fig.~\ref{fig:hq_shading} shows two color images that have been rendered using the one-pixel point rasterization technique described in this section.
We can see that with $\alpha = 0.01$ (right image) the aliasing is significantly reduced and for example the letters are easier to read.

\subsubsection*{Environment Map}
To represent the environment, we place a spherical environment map of resolution $1024^2$ around the scene.
Analog to the scene points, the environment map contains neural features, which are rendered as background.
Both, the features of the point cloud and those of the environment map are learned using derivatives computed in the following backward pass.

\subsubsection*{Backward Pass}
\label{sec:backward}

The backward pass of our point rasterizer first computes the partial derivatives of the render function \eqref{eq:render} w.r.t. its parameters.
\begin{equation}
	\frac{\delta \Phi}{\partial C}, 	\frac{\delta \Phi}{\partial R}, \dots
	\label{eq:partial}
\end{equation}
Using the chain rule, we can then compute the parameter's gradient w.r.t.~the loss and pass it to the optimizer.
The difficult part of this calculation are the partial derivatives of the structural parameters $C, R, t\ \text{and}\ x$. 
This will be explained in the following section.
The derivatives of the texture $\tau$ and environment map $E$ are straightforward and will not be detailed in this work.

\begin{figure}[]
	\hfil
	\includegraphics[width=0.5\linewidth]{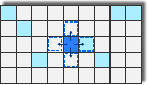}
	\caption{Pixel lookup for the spatial gradient computation of the center (dark blue) point. The white pixels are from the background and the teal pixels are other rasterized points.}
	\label{fig:difference}
\end{figure}

The problem of deriving $\Phi$ w.r.t. the structural parameters is the rounding operation Eq. \eqref{eq:round} of our one-pixel point rasterizer.
Due to the blending operation \eqref{eq:blend}, it is not possible to compute the change of a pixel's color by moving a point out of this pixel, without storing a list of points with each pixel and reevaluating the blending function.
So instead, we use an approximation that works well in practice.
The principle is shown in Fig.~\ref{fig:difference}. We virtually shift the projected point at $p = (u,v)$ one pixel in each direction and compute the change of intensity \textit{at the target pixels}, which is possible if we know the depth value at the target pixel.
The induced intensity change of these shifts are:
\newcommand\at[2]{\left.#1\right|_{#2}} 
\begin{equation}
\at{\frac{\Delta I}{\Delta u}}{p=(u+1,v)},\quad \at{\frac{\Delta I}{\Delta u}}{p=(u-1,v)}, \quad \dots
\label{eq:induce}
\end{equation}
from which we can approximate the gradient for $u$ (equally for $v$) at the desired location with 
\begin{equation}
\at{\frac{\partial I}{\partial u}}{p=(u,v)} \approx \frac{1}{2}\left({\at{\frac{\Delta I}{\Delta u}}{p=(u-1,v)} + \at{\frac{\Delta I}{\Delta u}}{p=(u+1,v)}} \right).
\label{eq:gradp}
\end{equation}
To compute each induced change of \eqref{eq:induce}, multiple cases have to be considered.
As shown in Fig.~\ref{fig:difference}, neighboring elements of $p$ can either show a color from the environment map (background) or from other rasterized points.
In the first case, the point simply overwrites the background, otherwise, three different cases have to be considered.
Firstly, the virtually shifted point can be completely behind the neighbor in which case no intensity change would be induced.
Secondly, the shifted point can be in front of the neighbor replacing the old color with the point's texture.
Lastly, the neighbor point can have a similar depth value according to the fuzzy depth test \eqref{eq:depthtest}, which causes us to have
	to compute the change of the blend function \eqref{eq:blend} if $p$ is added to the neighboring $\Lambda$.
In summary, the four cases to compute local spatial gradients are $(\text{with}\ (i,j) = (u \pm 1, v \pm 1))$:
%
\begin{align}
	&\at{\frac{\Delta I}{\Delta u}}{p=(i,j)} =
	\begin{cases}
		\tau(u,v) -I_l(i,j)                  									 & \Lambda_{l,u,v} = \emptyset \\
		0            & z > (1+\alpha)\text{min}_z(u,v)				\\
		\tau(u,v) - I_l(i,j) & z (1+\alpha) < \text{min}_z(u,v) \vspace{0.4em} \\
		\frac{ |\Lambda_{i,j}| I_l(i,j)+ \tau(u,v)}{1+|\Lambda_{i,j}|}  - I_l(i,j)    & \text{else} 
	\end{cases}
	\label{eq:spatialderiv}
\end{align}

\subsubsection*{Stochastic Point Discarding}
\label{sec:point_discarding}

Especially at small resolution layers, hundreds of points can pass the fuzzy depth-test of a single pixel, resulting in dramatic overdraw.
To reduce this number, we apply a stochastic point-discarding technique similar to \citet{enderton2010stochastic}.
In a preprocess, we compute the world-space radius $r_{world}$ as
the distance to the $4th$-nearest point. 
This radius is projected into screen-space during rendering, giving an approximation of the circular splat-size $r_{\mathbf screen}$.
Based on this size we then discard points stochastically to obtain a desired number of points per pixel
by assigning each point a uniform random value $\beta \in [0,1]$ and rendering it only if the following condition is met:
\begin{equation}
\frac{ r_{\mathbf screen} }{\sqrt{1-\beta}}  > \frac{1}{\gamma}
\label{eq:discard}
\end{equation}
The parameter $\gamma$ roughly controls how many points are rendered per pixel.
In our experiments, we use $\gamma=1.5$ for all datasets and scenes, see also Fig.~\ref{fig:abl_disc}.

\subsubsection*{Implementation Details}
\label{sec:implementation}
To implement the forward pass efficiently, we mainly follow the work of \citet{schutz2021rendering}.
They propose a three-pass point rasterizer that outperforms traditional point rendering with \texttt{GL\_POINTS}.
After a depth-only render pass, the color is accumulated for all points that pass the fuzzy depth-test.
During accumulation, we track the number of points per pixel so we can compute Eq.~\eqref{eq:blend} without writing the set $\Lambda_{u,v}$ to memory.
To further improve efficiency, we adopted the proposed blocked-morton-shuffle on the point cloud and use optimized local reductions inside the rasterization kernels.
A side-effect of not explicitly storing $\Lambda_{u,v}$ is that during backpropagation all points have to be rendered again.
However, this re-rendering only marginally impacts performance as the total backpropagation time is dominated by the neural networks.

To further improve the pose estimation of our inverse rendering pipeline, 
we follow state-of-the-art SLAM systems \cite{engel2017direct,mur2017orb,ruckert2021snake} that optimize the rigid transformation $(R,t) \in SE(3)$ in tangent-space.
Linearized pose increments are expressed as the Lie-algebra elements $x \in \mathfrak{se}(3)$ and applied to a transformation by
\begin{equation}
(R',t')= \exp(x) \cdot (R,t),
\label{eq:tangent}
\end{equation}
where $\exp(x)$ is the exponential map from $ \mathfrak{se}(3)$ to  $SE(3)$.
To implement tangent-space optimization in libTorch \cite{jia2014caffe}\cite{paszke2019pytorch}, we create a tangent tensor for the pose which is updated by the optimizer.
After each step, the tangent is applied to the original pose \eqref{eq:tangent} and reset to zero.

\subsection{Neural Renderer}
\label{sec:neural}

The neural renderer step (Fig.~\ref{fig:pipeline_overview} center) uses the multi-resolution neural images to produce a single HDR output image.
It consists of a four layer fully convolutional U-Net with skip-connections, where the lower resolution input images are concatenated to the intermediate feature tensors.
Downsampling is performed using average-pooling and the images are upsampled by bilinear interpolation.
As convolution primitives, we use \textit{gated convolutions}~\cite{yu2019free} which have been initially developed for hole-filling tasks and are therefore well suited for sparse point input.
Overall the network architecture is similar to \citet{aliev2020neural} with one less layer (see Sec.~\ref{sec:unet-layers})
and a few modifications to enable HDR imaging. 
First, we remove the batch-normalization layers, as they normalize the mean and standard deviation of intermediate images to fixed values.
The total sensor irradiance is therefore lost and cannot be propagated from a 3D point to the final image.
Usually, we store the neural point descriptors linearly, but if the scene's radiance range is sizable (larger than $1:400$) we use a logarithmic scaling, otherwise convergence issues with the optimizer can occur. 
For logarithmic descriptors, we convert them to linear space during rasterization so that the convolution operations only use linear brightness values.

\subsection{Differentiable Tone Mapping}
\label{sec:tonemapper}

The HDR output image of the neural renderer is converted to LDR by a learnable tone mapping operator (Fig.~\ref{fig:pipeline_overview} right), that mimics the physical lens and sensor properties of digital photo cameras.
The first tone mapping stage applies brightness correction to the HDR image $I_{HDR}$ using the estimated exposure value $\text{EV}_i$ of image~$i$.
\begin{equation}
	I_{e} = \frac{I_{HDR}}{ 2^{\text{EV}_i}}
\end{equation}
If image meta information is available, we initialize $\text{EV}_i$ using Eq.~\eqref{eq:ev},
where $f$ is the f-number of the optical system, $t$ the exposure time in seconds, $S$ the ISO arithmetic speed rating, and $\overline{\text{EV}}$ the mean exposure value of all images.
Otherwise, we initialize $\text{EV}_i$ to zero for all images.
\begin{equation}
	\text{EV}_i = \log_2 \left(  \frac{f_i^2}{t_i} \right) + \log_2 \left( \frac{S_i}{100} \right) - \overline{\text{EV}}
	\label{eq:ev}
\end{equation}

After brightness correction, we compensate for a changing white balance by estimating the white point $(R_i^w, G_i^w, B_i^w)$ for each image.
During optimization, we keep $G_i^w$ constant so that the white point cannot change the total image brightness.

As a next step, we model the \textit{vignette} effect of digital cameras, which is a radial intensity falloff due to various optical and sensor-specific effects.
The model we use \cite{goldman2010vignette} is a polynomial of the distance $r$ of a pixel $p$ to the vignette center $c_v$:
\begin{equation}
	I_v = I_w \cdot ( 1 + a_2 r^2 + a_4 r^4 + a_6 r^6)
\end{equation}
For better stability, we normalize $p$ and $c_v$ to be in the range $[0,1]$.
The coefficients $a_i$ are initialized to zero and $c_v$ is initialized to the image center.

\begin{table*}
	\begin{tabular}{@{}l|rcccccc@{}}
		\toprule
		Scene      & \#Points         & \#Images & Resolution & Exposure & White Balance & Camera Model    & Initial Reconstruction \\ \midrule
		M60        & 9,713,277		  & 313      & $2048\times1080$  & Constant   & Constant        & Pinhole + Dist. & COLMAP                 \\
		Train      & 11,818,812         & 301      & $1920\times1080$  & Auto     & Constant        & Pinhole + Dist. & COLMAP                 \\
		Playground & 12,521,941         & 307      & $1920\times1080$  & Auto     & Constant        & Pinhole + Dist. & COLMAP                 \\
		Lighthouse & 12,313,620         & 309      & $2048\times1080$  & Auto     & Constant        & Pinhole + Dist. & COLMAP                 \\
		Kemenate   & 34,144,401         & 405      & $2880\times1920$  & Constant   & Constant        & Pinhole + Dist. & Metashape              \\
		Boat       & 53,036,216         & 742      & $2880\times1920$  & Auto     & Constant        & Pinhole + Dist. & COLMAP                 \\
		Office     & 72,916,173         & 688      & $5472\times3648$  & Auto     & Auto          & Fisheye         & LiDAR + SLAM           \\ \bottomrule
	\end{tabular}
    \vspace{0.1cm}
	\caption{
		Overview of our evaluation scenes. 
		\textit{M60}, \textit{Train}, \textit{Playground}, and \textit{Lighthouse} are from the Tanks and Temples dataset.
		\textit{Kemenate} and \textit{Boat} were captured by the authors, \textit{Office} was provided by (anonymous).
	}
	\label{tab:scenes}
\end{table*}

The last stage in the tone mapping operator maps the linear RGB values to non-linear image intensities by applying the camera response function (CRF) \cite{grossberg2003determining} and an optional color space conversion from RGB to sRGB.
We combine both operations into a single function $R$, which is implemented using a 1D texture for every color channel.
\begin{equation}
	I_{ldr} = R(I_v)
\end{equation}
To guide the optimization towards a plausible response function, we initialize $R(x)$ to $x^{0.45}$ and add the following additional constraints based on \citet{debevec2008recovering}:
$ R(0) = 0, R(1)= 1, R''(x) = 0$.
These constraints ensure that the whole intensity range is covered and the response function is smooth.
Overexposed and underexposed pixels are clamped to the maximum and minimum output values of 1 and 0.
However, depending on the random network initialization, it is possible that the whole image is over/under exposed generating a zero gradient over the complete image.
Inspired by the LeakyReLU activation function \cite{xu2015empirical}, we define a separate response function $R_t$ during training that leaks small values instead of clamping them.
\begin{equation}
	R_{t}(x) =
	\begin{cases}
		\alpha x                            & x < 0         \\
		R(x)                                & 0\leq x\leq 1 \\

		\frac{-\alpha}{\sqrt{x}}+\alpha + 1 & 1 < x
	\end{cases}
	\label{eq:resonseleak}
\end{equation}
The last term of Eq.~\eqref{eq:resonseleak} asserts that the maximum leaked value is $(1+\alpha)$.
This is important in HDR scenes, because an overexposure of multiple magnitudes should not create a large gradient.
In practice we use $\alpha = 0.01$, which results in a maximum image intensity of $R_t(\infty) = 1.01$.


\section{Experiments}

\subsection{Datasets and Evaluation Methodology}
\label{sec:datasets}
We have collected seven datasets with differences in size, modality, and camera settings (see Tab.~\ref{tab:scenes}).
The first four scenes, i.e.~\textit{M60}, \textit{Train}, \textit{Playground}, and \textit{Lighthouse}, are from the popular tanks and temples dataset~\cite{knapitsch2017tanks}.
They consist of around 310 images in Full-HD resolution captured by a high-end video camera.
We use COLMAP \cite{schonberger2016structure} to reconstruct the initial dense point-cloud as well as the camera extrinsics and intrinsics.
The three remaining scenes, \textit{Kemenate}, \textit{Boat}, and \textit{Office}, are added to show how ADOP can handle different capture setups as well as to evaluate some specific aspects of our pipeline.
The \textit{Boat} scene was captured outdoors with a large variance in exposure time.
The \textit{Office} dataset consists of very high resolution fisheye input images and its point cloud was produced by a LiDAR scanner instead of a multi-view stereo approach.
For \textit{Kemenate}, the initial reconstruction was processed by Metashape~\cite{metashape}.
\textit{Please see also the accompanying video for results of all scenes.} 
	As shown in Table \ref{tab:scenes}, several scenes were captures with varying exposure, though only in \textit{Boat} the exact exposure value is stored in the meta data.
	In the other cases, we let ADOP estimate the real exposure value for each image.

For a quantitative evaluation we follow the standard approach in our community.
All input images are divided into a training set and a test set.
The training set is used to optimize all parameters of our rendering pipeline over multiple epochs.
We then let our system synthesize each test image and compare it to the real image using different metrics.
Since the test images are also prone to miscalibration, we do a realignment and exposure estimation of the test images after training has been completed.
For a fair comparison, we show the results with and without test refinement during our comparison to other approaches (see Tab.~\ref{tab:eval_tt}).

\begin{table}
	\ifx\arxivsubmission\undefined
	\small
	\fi
	\begin{tabular}{@{}l|r|r@{}}
		\toprule
		 		 & \multicolumn{1}{l|}{Inference (RTX 3080)} 	& Training  (A100)         \\ \midrule
		Nerf++		 & $\sim$ 183,000 ms                        		&$\sim$ 24 h        					 \\
		SVS  	  	 & $\sim$ 2,400 ms                        	& > 48 h          \\
		NPBG 	 	 & 50 ms                       		&$\sim$ \textbf{3 h}           \\
		ADOP (ours)	 & \textbf{27 ms}                  &$\sim$ 4 h  \\ \bottomrule
	\end{tabular}
	\centering
	    \vspace{0.1cm}
	\caption{Timings for novel view synthesis averaged over the four tanks and temples scenes, and approximate training times. 
	For SVS and Nerf++ we used the provided standard parameters for training.
	}
	\label{tab:timings_tt}
\end{table}

\begin{table*}
	\tabcolsep=0.15cm
	\begin{tabular}{@{}l|lr|lr|lr|lr|lr|lr|lr@{}}
		& \multicolumn{2}{c|}{Train}                   & \multicolumn{2}{c|}{Playground}              & \multicolumn{2}{c|}{M60}                     & \multicolumn{2}{c|}{Lighthouse}              & \multicolumn{2}{c|}{Kemenate}                & \multicolumn{2}{c|}{Boat}                    & \multicolumn{2}{c}{Office}                  \\
		& VGG $\downarrow$ & \multicolumn{1}{l|}{Diff} & VGG $\downarrow$ & \multicolumn{1}{l|}{Diff} & VGG $\downarrow$ & \multicolumn{1}{l|}{Diff} & VGG $\downarrow$ & \multicolumn{1}{l|}{Diff} & VGG $\downarrow$ & \multicolumn{1}{l|}{Diff} & VGG $\downarrow$ & \multicolumn{1}{l|}{Diff} & VGG $\downarrow$ & \multicolumn{1}{l}{Diff} \\ \midrule
		Baseline                                 & 514.3            &                           & 374.8            &                           & 369.5            &                           & 508.5            &                           & 204.7            &                           & 652.9            &                           & 186.8            &                          \\
		\ + Env.                    & 490.8            & -5\%                      & 374.8            & 0\%                       & 364.1            & -2\%                      & 440.1            & -14\%                     & 202.5            & -1\%                      & 646.1            & -1\%                      & 185.5            & -1\%                     \\
		\ + Env. + TM               & 412.0           & -19\%                     & 352.2            & -6\%                      & 360.3            & -3\%                      & 381.1            & -25\%                     & 189.3            & -8\%                      & 555.7            & -15\%                     & 145.4            & -22\%                    \\
		\ + Env. + TM + SO & \textbf{368.1}   & \textbf{-28\%}            & \textbf{318.7}   & \textbf{-15\%}            & \textbf{317.7}   & \textbf{-14\%}            & \textbf{350.0}     & \textbf{-31\%}            & \textbf{180.2}   & \textbf{-12\%}            & \textbf{535.6}   & \textbf{-18\%}            & \textbf{131.7}   & \textbf{-30\%}           \\ \bottomrule
	\end{tabular}
    \vspace{0.1cm}
\caption{Quantitative ablation study on the major contributions of ADOP.
The improvements due to the enviroment map (Env), the tone mapper (TM), and the structure optimization (SO) differs between the scenes.
For example, the M60 scene was captures with fixed exposure hence the tone mapper only has a small impact.
The structure optimization is beneficial in all cases. A qualitative evaluation for the Train scene is shown in Figure~\ref{fig:abl_modules}.
}
\label{tab:abl_modules}
\end{table*}

\subsection{Training Details and Timings}

Our inverse rendering pipeline (see Figure \ref{fig:pipeline_overview}) is optimized end-to-end using the C++ front-end of libTorch.
The neural network and the 4-element point descriptors are initialized randomly and optimized using ADAM \cite{kingma2014adam} with an initial learning rate of $0.0002$ and $0.08$, respectively.
If the radiance variation is high (see Sec.~\ref{sec:hdr_boar}), the descriptors are stored logarithmically and their learning rate is reduced by a factor of 10.
The remaining parameters, for example, camera pose and exposure time are updated using the SGD optimizer.
Over time, each learning rate is reduced when a plateau is detected.
For the exact values the reader is referred to the configuration files provided in the source code.
To further improve the robustness of our pipeline, we delay the optimization of structural parameters and photometric parameters for 25 epochs.
At that point, the rendered image is a blurry version of the target image and spatial gradients contain reasonable values.

The training for all scenes was done on an NVidia A100 with 40GB video ram. The batch size was set to 16 with a crop size of $512\times512$ pixels.
The four tanks and temples scenes were trained for 400 epochs, which takes around 3.5 hours per scene.
The remaining scenes were trained for 800 epochs.
Table \ref{tab:timings_tt} shows a inference and training time comparison of ADOP to Nerf++, SVS and NPBG.
Our method trains slightly slower than NPBG, mostly due to our more complex backpropagation, but still substantially faster than Nerf++ and SVS.
At inference time, our method is the fastest and achieves real-time performance (<33ms) on an NVidia RTX3080.

\subsection{Loss Function}
\begin{figure}
	\centering
	\includegraphics[width=\linewidth]{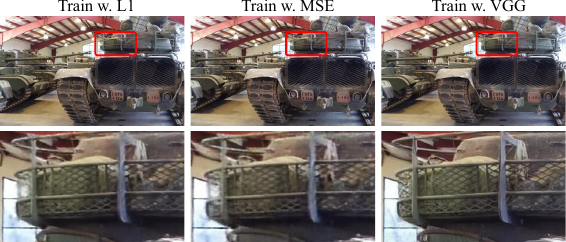}
	\vspace{0.05cm}
	\ifx\arxivsubmission\undefined
	\small
	\fi
	
	\centering
	\begin{tabular}{@{}l|rrr|rrr@{}}
		\toprule
		& L1 $\downarrow$            & MSE   $\downarrow$          & VGG   $\downarrow$          & PSNR   $\uparrow$        & LPIPS $\downarrow$         & SSIM  $\uparrow$         \\ \midrule
		Train w. L1  & \textbf{0.029} & \textbf{0.0033} & 487.43          & \textbf{25.10} & 0.264          & \textbf{0.846} \\
		Train w. MSE & 0.034          & 0.0037          & 558.84          & 24.67          & 0.322          & 0.822          \\
		Train w. VGG & 0.032          & 0.0035          & \textbf{357.08} & 24.79          & \textbf{0.145} & 0.804          \\ \bottomrule
	\end{tabular}
	\captionof{figure}{
		Effect of training ADOP with different loss functions.
		Using the VGG loss for training improves the image quality significantly even though some metrics favor a training with L1 loss.
	}
	\label{fig:abl_loss}
\end{figure}

\MS{candidate for appendix}
We have trained ADOP on the M60 scene using typical loss functions also used in prior work: L1, MSE, and the VGG perceptual loss \cite{johnson2016perceptual}.
For evaluation we use the LPIPS loss \cite{zhang2018perceptual}, the peak signal-to-noise ratio (PSNR), and the structured similarity index (SSIM).
The results are shown in Fig.~\ref{fig:abl_loss}.
Using the VGG loss, the most details are visible and the overall sharpness is significantly improved compared to the other losses.
All further experiments thus use VGG as loss function.

\subsection{Ablation Study}
\begin{figure}
	\centering
	\includegraphics[width=\linewidth]{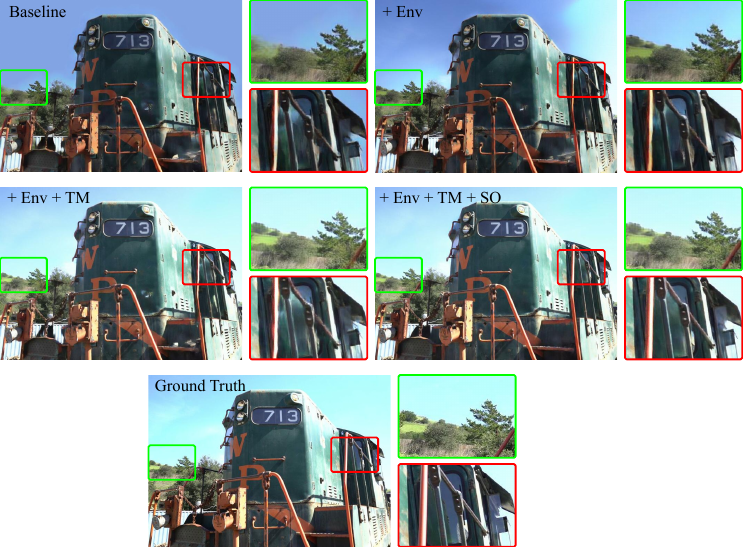}\vspace{0.1cm}
	\ifx\arxivsubmission\undefined
	\small
	\fi	
	\centering
	\captionof{figure}{
		Qualitative ablation study of the results presented in Table~\ref{tab:abl_modules}.
	}
	\label{fig:abl_modules}
\end{figure}

Compared to previous work of neural point-based rendering, we have proposed several ideas to improve the quality of novel view synthesis.
In Tab.~\ref{tab:abl_modules} and Fig.~\ref{fig:abl_modules}, we analyze how much each module contributes to the final result.
The first row shows a baseline experiment, which is similar to NPBG \cite{aliev2020neural}.
After that we enable one by one, the environment map (Env), the tone mapper (TM), and the structure optimization (SO).
In the latter, we initialize the scene using the COLMAP reconstruction and then optimize the camera pose, camera model, and point cloud during the training stage.
All together, a significant improvement in image quality is achieved with a VGG loss reduction between 12\% and 30\%.
The impact of the individual modules differs between the scenes.
The indoor scenes (M60, Kemenate, and Office) are not improved by the environment map. The tonemapper shows the largest impact on scenes with a high variance in exposure times (Train, Lighthouse, Boat, Office). 
However, structure optimization yields a positive change on all datasets.

\subsection{Robust Pose Correction}
\begin{figure}
	\centering
	\includegraphics[width=\linewidth]{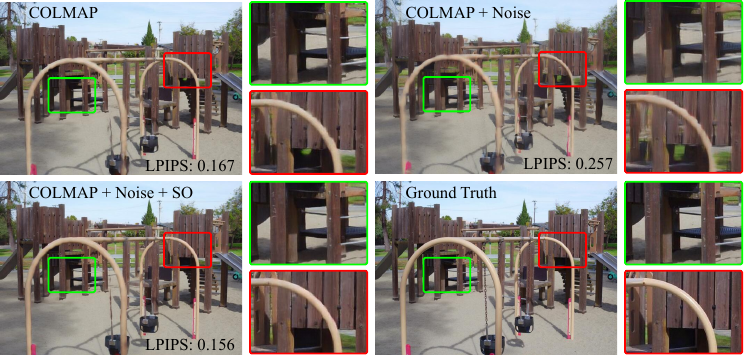}
	
	\vspace{0.1cm}
	\ifx\arxivsubmission\undefined
	\small
	\fi
	\begin{tikzpicture}
		\begin{axis}[
			height=5cm,
			width=0.90\linewidth,
			at={(0,0)},
			ymin=250,
			ymax=700,
			xmin=0,
			xmax=400,
			minor tick num =1,
			minor tick style={draw=none},
			ylabel=VGG (Train),
			xlabel=Epoch,
			legend style={font=\footnotesize},
			xlabel style={font=\footnotesize},
			ylabel style={font=\footnotesize},
			x tick label style={
				/pgf/number format/assume math mode, font=\tiny},
			y tick label style={
				/pgf/number format/assume math mode, font=\tiny},
			]
			\addplot[color=red,tension=0.7] table [x=ep, y=vgg, col sep=comma] {tables/ref_loss.txt};
			\addplot[color=blue,tension=0.7] table [x=ep, y=vgg, col sep=comma] {tables/no_op_loss.txt};
			\addplot[color=green,tension=0.7] table [x=ep, y=vgg, col sep=comma] {tables/op_loss.txt};			
			\legend{ COLMAP, COLMAP + Noise, COLMAP + Noise + SO}				
		\end{axis}
	\end{tikzpicture}
	\caption{
		Without structure optimization (SO), neural point-based rendering is sensitive to a bad initialization resulting in severe image artifacts (top right).
		However, our SO technique (bottom left) can recover from the erroneous input and outperforms the COLMAP initialization after 400 training epochs.
	}
	\label{fig:abl_pose}
\end{figure}
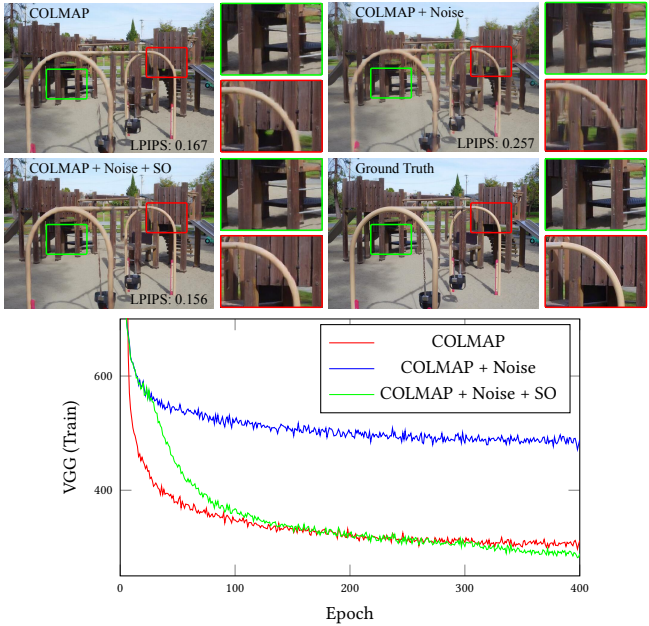

To further prove the robustness of our approximate gradient computation (see Sec.~\ref{sec:backward}), we add Gaussian noise to the camera position and rotation.
As shown in Fig.~\ref{fig:abl_pose}, ADOP is robust to a bad initial calibration and can also be used for image to point cloud alignment (see the supplemental material).

\subsection{U-Net Layers}\label{sec:unet-layers}
\begin{figure}
	\centering
	\includegraphics[width=\linewidth]{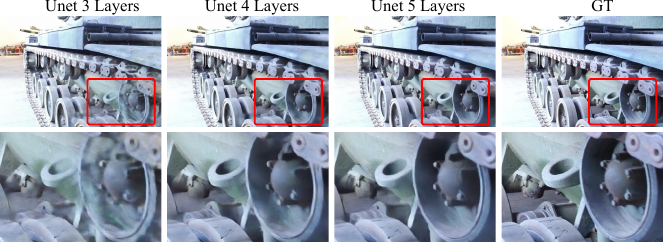}
	\vspace{0.05cm}
	\ifx\arxivsubmission\undefined
	\small
	\fi
	
	\centering
	\begin{tabular}{@{}l|r|r|rr|rr@{}}
		& \multicolumn{1}{l|}{}         & Train Loss     & \multicolumn{2}{c|}{Test Loss}   & \multicolumn{2}{c}{Time (ms)} \\
		& \multicolumn{1}{l|}{\#Params} & VGG $\downarrow$  & VGG $\downarrow$  & LPIPS $\downarrow$ & FP32             & FP16            \\ \midrule
		3 Layers & 136,387                        & 302.5          & 330.9          & 0.1364          & \textbf{34.264}  & \textbf{22.604} \\
		4 Layers & 574,651                        & 262.8          & \textbf{317.0} & \textbf{0.1191} & 41.343           & 26.665          \\
		5 Layers & 2,335,923                       & \textbf{245.7} & 325.8          & 0.1239          & 48.106           & 30.798          \\ \bottomrule
	\end{tabular}
	\captionof{figure}{
		Quality comparison on the number of Unet-layers inside the neural renderer.
		More layers improves the hole-filling capabilities (see images) but also reduces rendering efficiency.
		On some scenes, a 5 layer network is also prone to overfit on the training images (low train loss, but high test loss).
	}
	\label{fig:abl_layers}
\end{figure}

Due to the improved consistency of the input, we can afford to use a smaller and faster neural network for final reconstruction.
While other methods often use a five layer network \cite{aliev2020neural,thies2019deferred}, we show in Fig.~\ref{fig:abl_layers} that also fewer layers can produce good results.
Note that with the number of layers in particular the hole-filling capabilities decrease, so that the chosen number of layers should be selected according to the point densities.
In our experiments, we found that a four layer network improves efficiency and is less prone to overfitting, all shown results thus were generated with four layers.

\subsection{Stochastic Point Discarding}
\begin{figure}
	\centering
	\includegraphics[width=.98\linewidth]{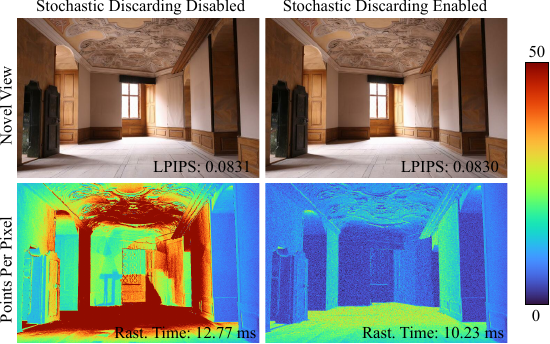}
	\captionof{figure}{
		Stochastic discarding roughly halves the number of blended points, which increaes rasterization efficiency by 15-25\%, without impact on quality.
	}
	\label{fig:abl_disc}
\end{figure}

In Sec.~\ref{sec:point_discarding} we proposed a stochastic point discarding technique to reduce the pixel overdraw.
To examine a possible quality loss, we train ADOP with and without point discarding on the \textit{Kemenate} dataset. 
We measure the number of blending operations, render time, and image quality, the result is presented in Fig.~\ref{fig:abl_disc}.
Visually, no difference is noticeable, which is backed by the perceptual rendering loss.
However, on average only 50\% of the points are blended, which reduces rendering time by 20\%.

\begin{table*}
	\centering
	\includegraphics[width=\linewidth]{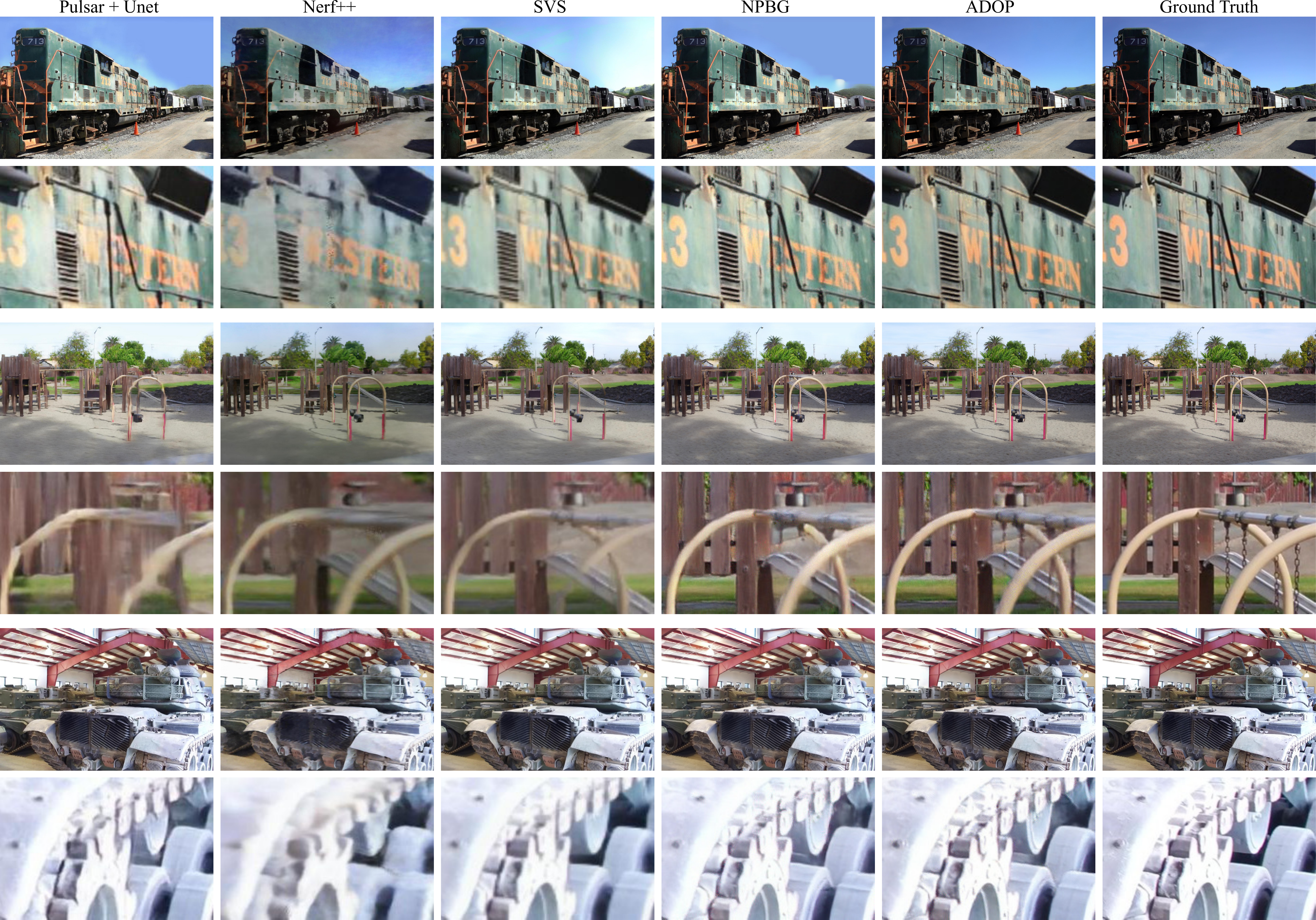}
	\captionof{figure}{
		Comparative results of novel-view synthesis on the \textit{Train}, \textit{Playground}, and \textit{M60} scene.
		The table below provides a quantitative comparision, in this figure we show images for the methods that perform best.
		Note that due to memory constraints SVS has been evaluated in half-resolution.
	}
	\label{fig:eval_tt}
	\vspace{0.4cm}	
	\ifx\arxivsubmission\undefined
	\small
	\fi
	\begin{tabular}{@{}l|rrr|rrr|rrr|rrr@{}}
		\toprule
		& \multicolumn{3}{c|}{Train}                              & \multicolumn{3}{c|}{Playground}                           & \multicolumn{3}{c|}{M60}                                 & \multicolumn{3}{c}{Lighthouse}                            \\
		Method        & VGG $\downarrow$ & LPIPS $\downarrow$ & PSNR $\uparrow$ & VGG  $\downarrow$ & LPIPS  $\downarrow$ & PSNR $\uparrow$ & VGG $\downarrow$ & LPIPS $\downarrow$ & PSNR  $\uparrow$ & VGG  $\downarrow$ & LPIPS  $\downarrow$ & PSNR $\uparrow$ \\ \midrule
		Synsin + Unet & 706.0            & 0.3853             & 16.97           & 521.3             & 0.3198              & 21.81           & 564.3            & 0.2904             & 20.16            & 679.7             & 0.3655              & 15.41           \\
		Pulsar + Unet & 677.9            & 0.3418             & 17.78           & 661.2             & 0.3849              & 20.00           & 508.2            & 0.2403             & 21.42            & 587.1             & 0.3112              & 17.82           \\
		NRW           & 817.5            & 0.4552             & 14.44           & 632.8             & 0.4110              & 19.47           & 741.2            & 0.4476             & 16.96            & 709.5             & 0.3835              & 14.88           \\
		NeRF++           & 857.5            & 0.5168             & 18.04           & 696.8             & 0.5292              & 22.24           & 700.5            & 0.4378             & 23.06            & 741.2             & 0.4609             & 20.06           \\
		SVS  (half res.)          & 633.1           & 0.323              & 17.43           & 516.6             & 0.3462              & 22.11           & 461.1            & 0.2440             & 23.74            & 606.5             & 0.3232              & 17.12           \\
		NPBG          & 521.9            & 0.2094             & 17.66           & 389.9             & 0.1816              & 23.31           & 380.69           & 0.1494             & 24.15            & 500.0             & 0.2140              & 17.54           \\
		ADOP   & \textbf{422.6} &\textbf {0.1603}    & \textbf{21.62}  &\textbf {345.9}    & \textbf{0.1594}     & \textbf{25.00}  & \textbf{342.8}   & \textbf{0.1344}    & \textbf{25.07}   & \textbf{375.1}    & \textbf{0.1536}     & \textbf{22.44}  \\  \midrule
		ADOP w. TR  & {368.1} & {0.1438}    & {23.19}  & {318.7}    & {0.1464}     & {25.78}  & {317.7}   & {0.1189}    & {25.84}   & {350.0}    & {0.1439}     & {23.04}  \\ \bottomrule
	\end{tabular}
	\centering
	\captionof{table}{
		Quantitative evaluation of novel view synthesis on the four scenes of the tanks and temples dataset.
		The values in this table represent the mean loss over all test images.
		ADOP outperforms the other approach with and without test refinement (TR).
		In the latter, we estimate the expsoure value and refine the camera pose of the test images.
	}
	\label{tab:eval_tt}
\end{table*}

\subsection{Comparison With Prior Work}
We evaluate our system with state-of-the-art prior work on the scenes from the tanks and temples dataset~\cite{knapitsch2017tanks}, see~Tab.~\ref{tab:scenes}.
The methods we compare against are SynSin \cite{wiles2020synsin}, Pulsar \cite{Lassner_2021_CVPR}, Neural Rendering in the Wild (NRW) \cite{meshry2019neural}, NeRF++ \cite{zhang2020nerfpp}, Stable View Synthesis (SVS) \cite{riegler2021stable}, and Neural Point Based Graphics (NPBG) \cite{aliev2020neural}.
The first two are general differentiable rendering front-ends that have been adapted by us to the novel view synthesis problem.
They can be seen as similar to our approach and NPBG with the difference of using sphere and splat-based rendering instead of multi-resolution one-pixel point rendering.
	We did not compare against all related inverse rendering systems, because some are not applicable to our data \cite{loper2014opendr} and others are outperformed by Pulsar and Synsin \cite{liu2019soft,kato2020differentiable,lin2018learning} .

Pulsar, Synsin, Nerf++ and SVS only support pinhole cameras.
Therefore we train them on the undistorted images and camera parameters provided by COLMAP.
During evaluation the synthesized undistorted images are distorted again and compared to the ground truth.
NPBG, NRW, and our approach use one-pixel point rendering, which natively supports the camera models, so they have been trained on the original distorted images.

The results (see Fig.~\ref{fig:eval_tt}) show that ADOP is able to produce convincing results on all tested scenes.
Compared to the other methods, our renderings show less artifacts as well as preserve color, brightness, and sharpness better.
For example, in the close-up view of the playground scene, the metal bars of the swing are most detailed in the ADOP rendering.
In the M60 dataset, NPBG has similar visual results, but the quantitative evaluation (in Tab.~\ref{tab:eval_tt}) shows that our method still has the edge.
In the last row of Tab.~\ref{tab:eval_tt} we show the quantitative results of ADOP with test refinement (TR) enabled. TR compensates errors of the initial reconstruction by estimating the exposure value and camera pose of the test frames once training has been completed.
This further improves the loss however note that even without TR our approach outperforms the state-of-the-art.

\subsection{Fisheye Camera}
\begin{figure}
	\centering
	\includegraphics[width=\linewidth]{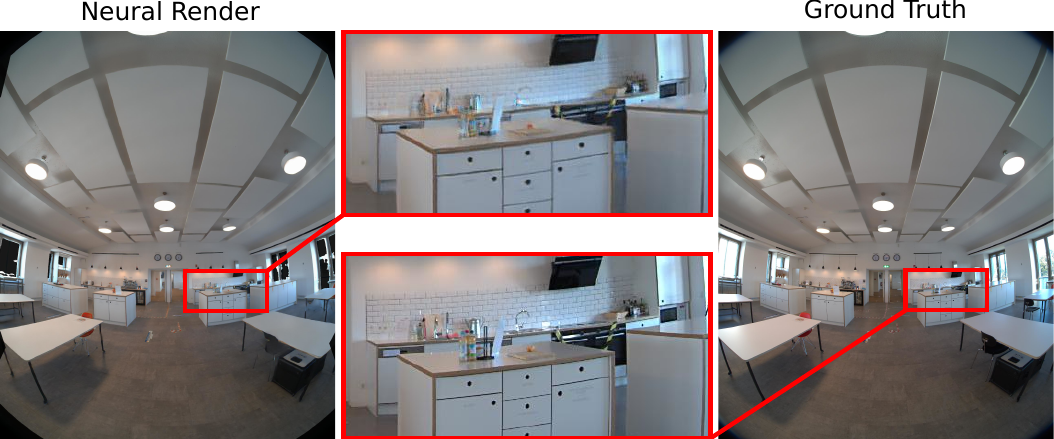}
	\captionof{figure}{
		ADOP is able to render fisheye camera systems at high quality.
	}
	\label{fig:fisheye_nav}	
\end{figure}

Our system is able to learn with images originating from fisheye cameras without needing to undistort the projection, resulting in high quality neural fisheye renders (see Fig.~\ref{fig:fisheye_nav}).
The office scene uses a LiDAR and SLAM system for poses and the point cloud, which can be noisy. As presented before, ADOP is able to correct this efficiently.

\subsection{View Extrapolation}
\begin{figure}
	\centering
	\includegraphics[width=\linewidth]{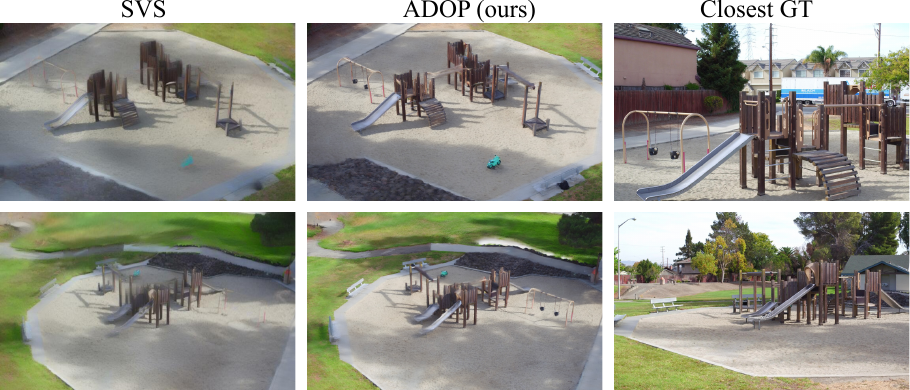}
	\captionof{figure}{
		View extrapolation with a large distance to the closest training image.
	}
	\label{fig:extrapolation}
	
\end{figure}

In Fig.~\ref{fig:extrapolation} we show reconstructions from a view that is far from the input views, comparing the results of our system with SVS.
In general, ADOP generates relatively good results in such cases compared to established methods, although this setup is generally difficult.

\subsection{HDR Neural Rendering}\label{sec:hdr_boar}

\begin{figure}
	\hfil
	\includegraphics[width=\linewidth]{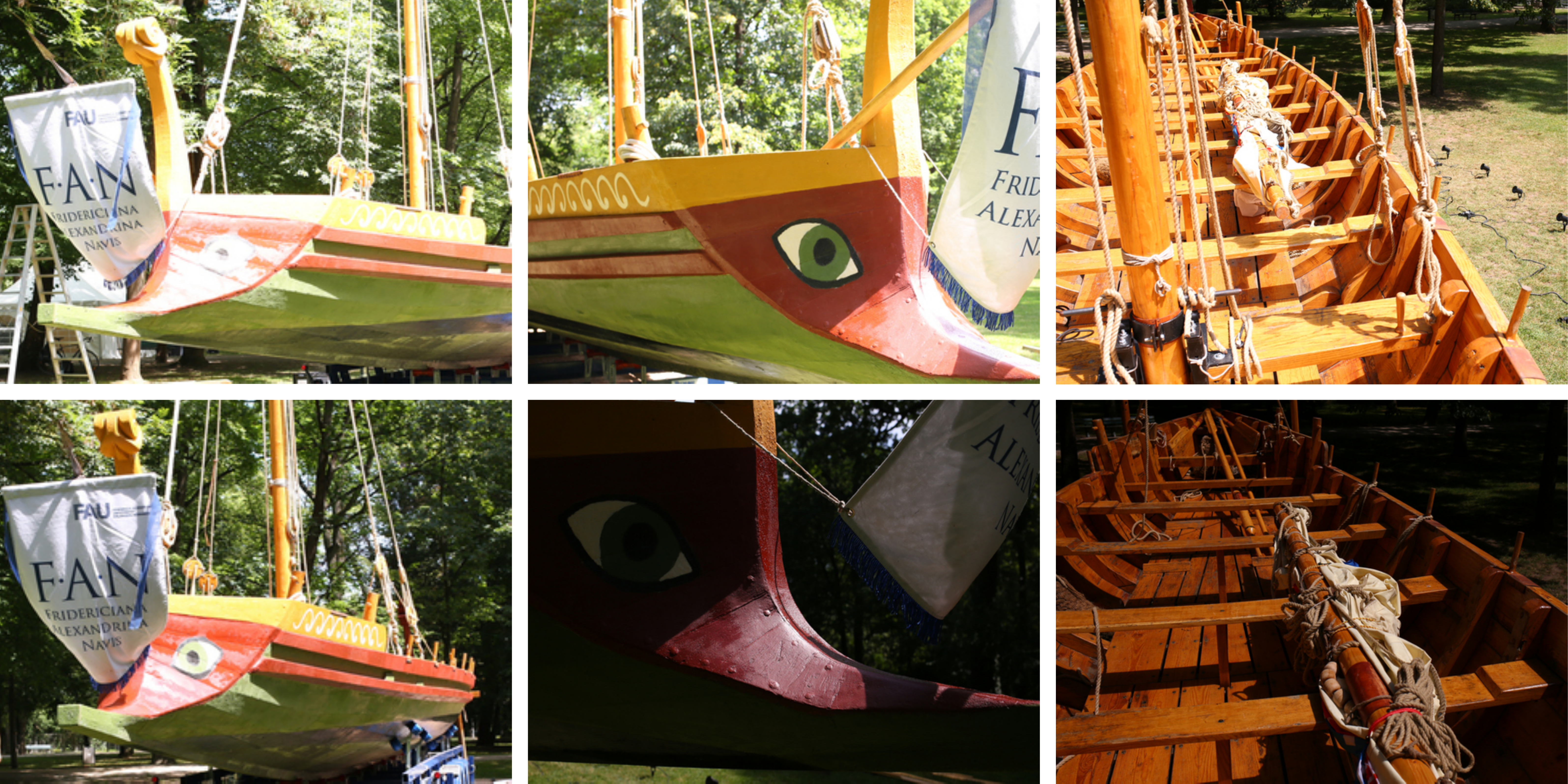}
	\centering
	\begin{tikzpicture}
	\begin{axis}[
	height=3cm,
	width=\linewidth,
	at={(0,0)},
	ymin=8,
	ymax=20,
	xmin=0,
	xmax=741,
	minor tick num =1,
	minor tick style={draw=none},
	ylabel=EV,
	ylabel style={font=\footnotesize},
	x tick label style={
		/pgf/number format/assume math mode, font=\tiny},
	y tick label style={
		/pgf/number format/assume math mode, font=\tiny},
	]
	\addplot[color=green,tension=0.7] table [x=frame, y=ev_abs, col sep=comma] {tables/ev.csv};
	\end{axis}
	\end{tikzpicture}
	\caption{
		Sample images of the boat dataset, taken with auto exposure.
		The exposure value (EV) for each image in the dataset is plotted below the images.
	}
	\label{fig:boot_input}
\end{figure}
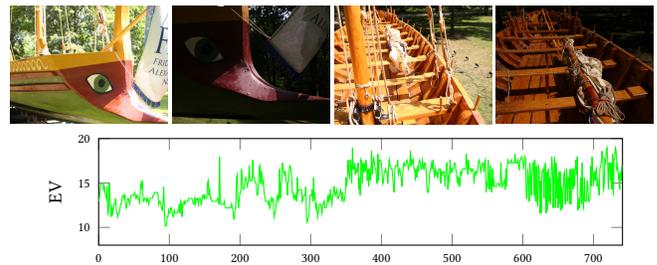

\begin{figure}
	\hfil
	\includegraphics[width=\linewidth]{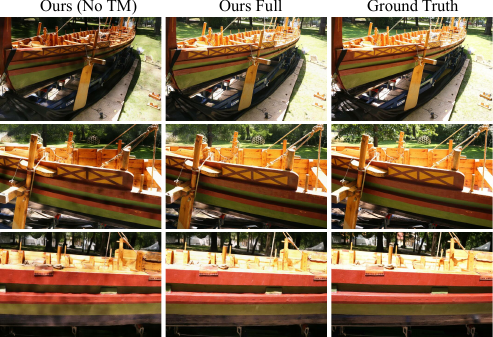}
	\caption{
		Novel views synthesized on the boat dataset.
		Left: no tone mapping and exposure correction results in splotchy artifacts, Center: HDR reconstruction with tone mapping and same exposure as reference photograph. Right: ground truth.
	}
	\label{fig:boot_synth}
\end{figure}
When capturing outdoor scenes, it is almost unavoidable to use varying exposure to be able to adapt to the huge range of brightness values.
To test the performance of our pipeline on such HDR scenes, we apply it to our \textit{Boat} dataset.
The camera was set to auto exposure, all other image related settings, such as aperture, ISO-Speed, and white balance were set constant for all frames.
In Fig.~\ref{fig:boot_input}, four samples from the dataset are shown, underneath the exposure value is plotted for every frame in the dataset.
The difference between the smallest and largest EV is $8.7$, which corresponds to a factor of $2^{8.7} =426.67$. 

Novel view synthesis results of the Boat scene are shown in Fig.~\ref{fig:boot_synth}.
For the images in the left column, uniform exposure was assumed, in the middle column the exposure value used by the tonemapper has been set to the real EV from the EXIF meta data and kept constant during training. 
Our system is then able to handle the high dynamic range and avoids splotchy artifacts.

\begin{figure}
	\hfil
	\includegraphics[width=.8\linewidth]{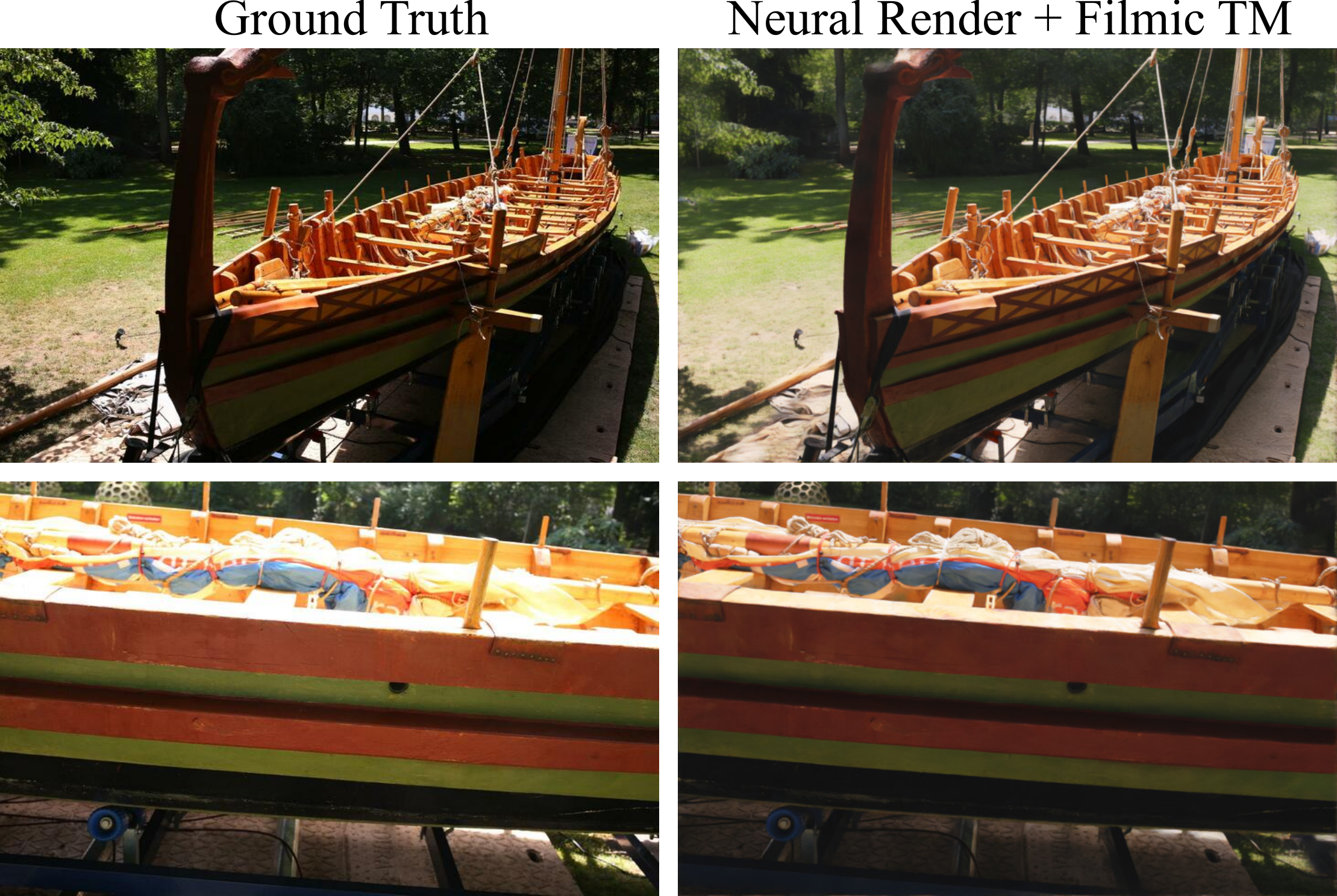}
	\caption{
	At inference time, we can replace the learned tone mapper (TM) by a filmic TM, which renders the images in a more natural look.
	}
	\label{fig:boot_output}
\end{figure}

Additionally, we can replace the learned TM at inference time, for instance by a filmic tone mapper \cite{uc2filmic}, that better resembles human perception.
The result is shown in Fig.~\ref{fig:boot_output}.
With the filmic tone mapper, the dark areas have significantly more contrast without overexposing the bright wood inside the boat.
The filmic TM also slightly reduces color saturation, as such colors look more natural.

\section{Limitations}

Due to the vast amount of different parameters, the search of suitable hyper parameters is non-trivial.
We have to balance learning rates of the texture color, structural parameters, tone mapping settings, and neural network weights.
An extensive grid-search was necessary to find viable settings that work well for all of our scenes.

Furthermore, the optimization of point position is not stable for moderate to large learning rates.
Our pipeline therefore requires a reasonable initial point cloud, for example, by a multi view stereo system or a LiDAR scanner.
We believe that this problem is caused by the gradient approximation during rasterization. 
It works well for camera model and pose optimization because the spatial gradient of thousands of points are averaged in one optimizer step.
For the positional point-gradients however, only a single approximate gradient is used to update its coordinates.
A very low learning rate is therefore required to average the point gradient over time.

Finally, due to the one-pixel point rendering, holes appear when the camera is moved too close to an object or the point cloud is very sparse, because the neural network architecture can only fill holes up to a certain size threshold.
When the camera moves, also flickering becomes noticeable in such situations--the effect is visible in parts of the accompanying video.
Using a deeper neural renderer network helps in these cases, at the price of reduced rendering performance.
Future work should be conducted here, for example one could try to dynamically generate new points during magnification that have interpolated neural descriptors or add a temporal component to the neural rendering.

\section{Conclusion}

We have presented ADOP, a novel fully differentiable neural point-based rendering pipeline that uses a set of calibrated images and an estimated point cloud of the scene as input.
%
All stages are differentiable, so we can refine all input parameters to generate an improved output.
By also including a photometric sensor model, we can handle input images with varying exposure and generate high dynamic range output.

Our experiments show that, due to the optimized input, we achieve superior rendering quality, especially for novel view synthesis.
At the same time, the one-pixel point renderer is very fast.
In combination with the fact that the improved consistency in the input allows us to use a simpler and thus faster neural network for reconstruction, we achieve real-time frame rates on a standard GPU, even for complex scenes.

\begin{center}
{\color{blue}\url{https://github.com/darglein/ADOP}}
\end{center}

\begin{acks}
Linus Franke was supported by the Bayerische Forschungsstiftung (Bavarian Research Foundation) AZ-1422-20.
%
%
%
%
\end{acks}

\bibliographystyle{ACM-Reference-Format}
\bibliography{common/0_Literature}

\appendix

\section{Image to Point Cloud Alignment}
\label{sec:image2points}

\begin{figure}
	\includegraphics[width=\linewidth]{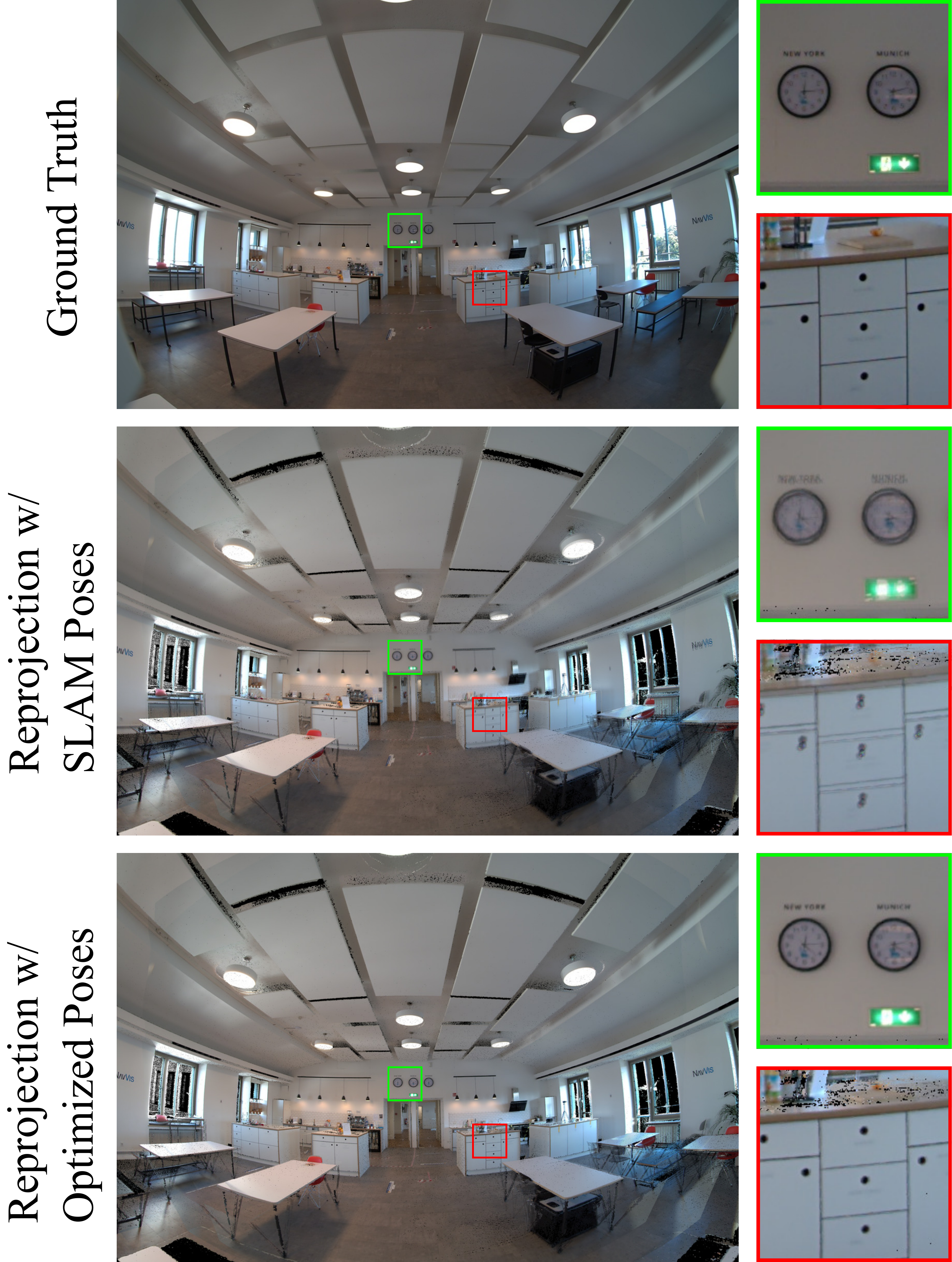}
	\caption{The initial camera pose estimates of the SLAM-System are slightly misaligned w.r.t. the LiDAR point cloud. Reprojecting the pixel color of several source views into a target view produces ghosting artifacts (center row).
		Our system is able to optimize the camera poses resulting in almost pixel perfect reprojections (bottom row).
	}
	\label{fig:pose_alignment2}
\end{figure}

Another application of our differentiable rendering pipeline is the alignment of camera images to point-clouds that were reconstructed by external devices.
(Anonymous) provided us a dataset captured by the (Anonymous) mobile scanning platform. 
This platform consists of a high performance LiDAR scanner and four $5472 \times 3648$ pixel fisheye cameras.
Their reconstruction software is able to combine multiple laser scans into a consistent point cloud and provides camera pose estimates for panorama generation and point cloud coloring.
However, the image to point cloud registration is not perfect.
Small errors during the SLAM-based tracking and vibrations due to the hand-held operation result in pose errors in the scale of millimeters.
If we then use these poses for precise operation, such as reprojecting the color of neighboring source images into a target view, small ghosting artifacts can be observed (see Figure \ref{fig:pose_alignment2} center row).
We train our neural rendering pipeline on this office dataset, which is composed of 688 images and 73M points, to synthesize all captured views.
During training, our system also optimizes the camera pose of each frame.
The refined poses are then used to reproject the colors into the same target frame as before (see Figure \ref{fig:pose_alignment2} bottom row).
It can be seen that the ghosting artifacts are mostly eliminated and the synthesized image is a lot sharper.
This experiment shows that the proposed method is able to perform pixel-perfect alignment of fisheye camera images to a point cloud of a LiDAR scanner.
To our knowledge, no other available differentiable renderer can fulfill this task, as they assume a pinhole camera model or are not able to handle a point cloud with 73M points.

\begin{table*}[]
	\centering
	\ifx\arxivsubmission\undefined
	\small
	\fi
	\begin{tabular}{@{}l|lc|rr|rr|rr|rr@{}}
		\toprule
		& Geometry       & \# Layers & Forward        & Backward        & Forward        & Backward        & Forward        & Backward        & Forward        & Backward        \\ \midrule
		Synsin             & Disc Splats   & 1      & 856.63         & 14.14           & 1692.63        & 17.15           & 3859.1         & 31.16           & 7342.05        & 25.73           \\
		Pulsar ($r=0.005$)             & Spheres        & 1      & 31.20          & 3.62             & 59.69         & 6.97            & 129.60         & 15.06           & 216.07        & 16.32            \\
			Pulsar ($r=0.02$)             & Spheres        & 1      & 36.81          & 3.02             & 70.78          & 5.80            & 152.89         & 12.53           & 263.59         & 14.91            \\
			Pulsar ($r=0.05$)             & Spheres        & 1      & 53.29          & 3.09             & 102.48          & 5.95            & 221.35         & 12.85           & 379.06         & 15.19            \\
		GL\_POINTS         & 1-Pixel Points & 1      & 0.3            & $\times$        & 0.54           & $\times$        & 1.09           & $\times$        & 1.85           & $\times$        \\
		Ours               & 1-Pixel Points & 1      & 0.83           & 0.62            & 1.09           & 0.78            & 1.59           & 1.39            & 2.33           & 2.34            \\
		Ours               & 1-Pixel Points & 4      & 1.4            & 0.77            & 1.82           & 1.24            & 2.52           & 1.74            & 3.65           & 2.8             \\
		\textbf{Ours + Stoc. Disc.} & \textbf{1-Pixel Points} & \textbf{4} & \textbf{1.28}   & \textbf{0.66}  & \textbf{1.66}   & \textbf{1.07}  & \textbf{2.15}   & \textbf{1.66}  & \textbf{3.08}   & \textbf{2.64}  \\ \midrule
		& \# Points      &        & \multicolumn{2}{c|}{1,348,406} & \multicolumn{2}{c|}{2,570,810} & \multicolumn{2}{c|}{5,400,615} & \multicolumn{2}{c}{10,283,243} \\ \bottomrule
	\end{tabular}
	\caption{
		Forward and backward render-time in milliseconds of a $1920\times 1080$ image on a RTX 2080 Ti.
		In comparison to other differentiable renderers, our approach is around two magnitudes more efficient.
		The highlighted row is our approach with stochastic point discarding (see Section \ref{sec:implementation}) that we use for scene refinement and novel-view synthesis. 
	}
	\label{tab:runtime}
\end{table*}

\section{Runtime Performance}
\label{sec:runtime}

Runtime performance has been a limiting factor for differentiable rendering systems in the past \cite{kato2020differentiable}.
Most software rasterization techniques exceed the \SI{100}{\milli\second} barrier even for small scenes and render resolutions \cite{Lassner_2021_CVPR}.
This limits their usefulness in real-world applications such as 3D reconstruction from high-resolution photographs.
Currently, the two most performant differentiable rendering methods that are able to process point-cloud data are Synsin~\cite{wiles2020synsin} and Pulsar~\cite{Lassner_2021_CVPR}.
Synsin, which is the default point render engine of PyTorch3D \cite{ravi2020accelerating}, splats each point to a disk and blends the $K$ nearest points of each pixel into the output image.
Pulsar converts each point to sphere and blends them with a similar approach as Soft Rasterizer~\cite{liu2019soft}.
Both methods are fully differentiable, meaning that the point position and color can be optimized during rendering.

Table \ref{tab:runtime} shows the measured GPU frame-time for Synsin, Pulsar, our approach, and OpenGL's default point rendering with GL\_POINTS.
These timings only include the rasterization itself without the neural network and tonemapper described in the previous sections.
For our method, we also include the rendering time of four layers in different resolutions.
This is a more fair comparison to the other methods because all four layers are required for the neural rendering network.
The output of Synsin and Pulsar is more complete and therefore a single layer can already be successfully used.
The right most columns of Table \ref{tab:runtime} show the forward and backward render time of a $1920\times 1080$ image for a point cloud with around $10$M points.
Both Synsin and Pulsar are not real-time capable at such dimensions with forward timings of  \SI{7342}{\milli\second} and  \SI{209}{\milli\second} respectively.
Our approach takes two magnitudes less time than Pulsar with a combined rendering time of \SI{3.65}{\milli\second} for all four layers.
This result is expected though because previous work has shown that software one-pixel point rendering can outperform hardware rasterization techniques \cite{schutz2021rendering}.
Point splatting and sphere rendering is inherently more complex because each point effects multiple output pixels.

If we enable stochastic discarding (see Section \ref{sec:implementation}), the rendering performance is further increased.
The largest gains are achieved for multi-resolution rendering of large point clouds.
For example, generating a multi-layer image of the cloud with 10.4M points takes 19\% less time with our stochastic discarding approach.
However, if only a single image layer is required, the speedup due to stochastic discarding reduces to 3\%, as points are only discarded if they fill less than one pixel in the output image.
The pixels inside the low resolution layers are much larger and therefore more points are discarded.
In comparison to native  GL\_POINTS rendering, our approach is only slightly slower (by about 26\%).
This is an impressive result, because we have implemented a three-pass blending approach with fuzzy depth-test as described in Section \ref{sec:implementation}.
The GL\_POINTS reference implementation in Table \ref{tab:runtime} uses a single pass without blending and standard GL\_LESS depth test.

\end{document}